\documentclass[a4paper,conference]{IEEEtran}

\usepackage{booktabs}       
\usepackage[hidelinks]{hyperref}

\ifCLASSINFOpdf
  \usepackage[pdftex]{graphicx}

\else

\fi

\ifCLASSOPTIONcompsoc
 \usepackage[caption=false,font=normalsize,labelfont=sf,textfont=sf]{subfig}
\else
 \usepackage[caption=false,font=footnotesize]{subfig}
\fi

\usepackage[table]{xcolor}
\usepackage{graphicx}
\usepackage{multirow}

\makeatletter
\global\let\oriCT@@do@color\CT@@do@color 

\begin{document}

\title{Classification of Bark Beetle-Induced Forest Tree Mortality Using Deep Learning}

\author{\IEEEauthorblockN{
Rudraksh Kapil\IEEEauthorrefmark{1}\IEEEauthorrefmark{2}\textsuperscript{\textsection},
Seyed Mojtaba Marvasti-Zadeh\IEEEauthorrefmark{2}\textsuperscript{\textsection}, 
Devin Goodsman\IEEEauthorrefmark{3},
Nilanjan Ray\IEEEauthorrefmark{1}, and
Nadir Erbilgin\IEEEauthorrefmark{2}
}
\IEEEauthorblockA{
\IEEEauthorrefmark{1}Department of Computing Science, University of Alberta \\
\IEEEauthorrefmark{2}Department of Renewable Resources, University of Alberta \\
\IEEEauthorrefmark{3}Canadian Forest Service, Natural Resources Canada, Edmonton, Alberta
}
Emails: \{rkapil, seyedmoj, nray1, erbilgin\}@ualberta.ca, devin.goodsman@nrcan-rncan.gc.ca
}
\maketitle
\begingroup\renewcommand\thefootnote{\textsection}
\footnotetext{Equal contribution.}
\endgroup


\begin{abstract}

Bark beetle outbreaks can dramatically impact forest ecosystems and services around the world. For the development of effective forest policies and management plans, the early detection of infested trees is essential. Despite the visual symptoms of bark beetle infestation, this task remains challenging, considering overlapping tree crowns and non-homogeneity in crown foliage discoloration. In this work, a deep learning-based method is proposed to effectively classify different stages of bark beetle attacks at the individual tree level. 
\textcolor{black}{The proposed method uses RetinaNet architecture (exploiting a robust feature extraction backbone pre-trained for tree crown detection) to train a shallow subnetwork for classifying the different attack stages of images captured by unmanned aerial vehicles (UAVs).}
Moreover, various data augmentation strategies are examined to address the class imbalance problem, and consequently, the affine transformation is selected to be the most effective one for this purpose. Experimental evaluations demonstrate the effectiveness of the proposed method by achieving an average accuracy of 98.95\%, considerably outperforming the baseline method by $\sim$10\%.
The code \textcolor{black}{and results are} publicly available at \href{https://github.com/rudrakshkapil09/BarkBeetle-Damage-Classification-DL}{https://github.com/rudrakshkapil09/BarkBeetle-Damage-Classification-DL}
\end{abstract}

\IEEEpeerreviewmaketitle

\section{Introduction}
\begin{figure}[t]
\centering
\includegraphics[width=0.98\linewidth]{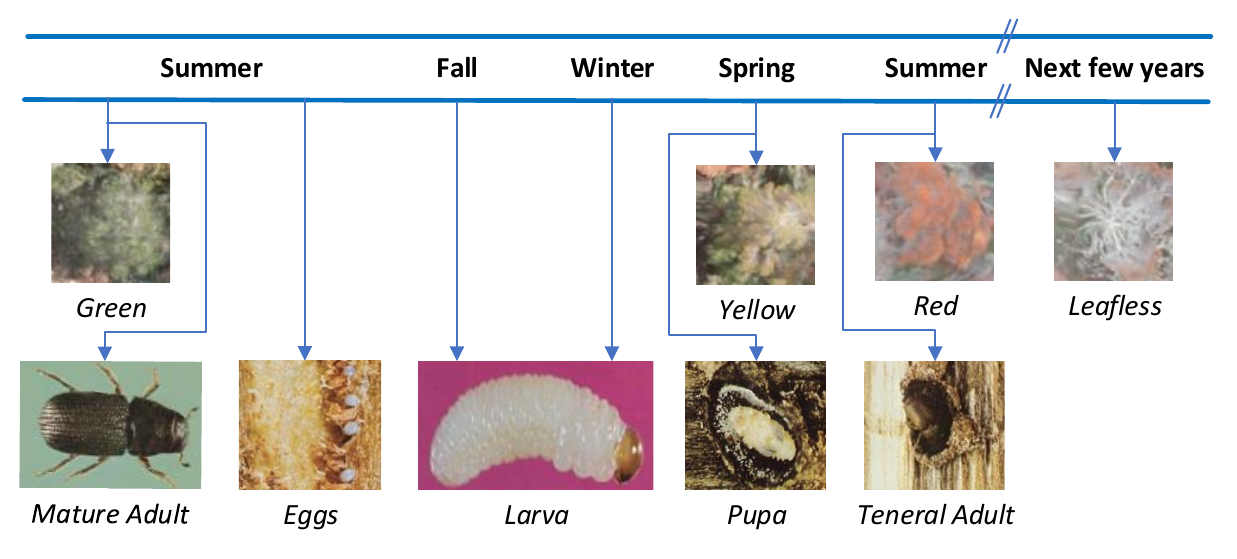}
\vspace{-.3cm}
\caption{Typical life cycle of bark beetles and their effect on host tree foliage over time. Beetle images have been adapted from \cite{13_safranyik_2006}.}
\label{fig:lifeCycle}
\vspace{-0.5cm}
\end{figure}

\noindent Bark beetle outbreaks significantly impact forests worldwide, 
thereby disrupting the functioning and properties of natural ecosystems.
As a result of various factors (e.g., population density, tree moisture \& condition, beetle \& host tree species), a successful bark beetle attack gradually reveals itself by affecting various parts of the host tree \cite{erbilgin_1}. Over time, the crown of an infested tree begins to fade -- there is a gradual change in foliage color from a healthy green to yellow, red, and finally a leafless (i.e., needle-less) gray. These are referred to as different attack stages). 
The rate of discoloration depends on the progress of bark-beetle induced fungal infection that interrupts nutrient and water flow through the phloem and xylem, as well as environmental conditions such as soil moisture content \cite{niemann_2005_discoloration}. The fading process is linked to the ecology of bark beetles (see Fig.~\ref{fig:lifeCycle}), in which female bark beetles bore tunnels (called oviposition galleries) in the phloem to lay their eggs, and the larvae hatch and excavate additional larval galleries to feed on phloem tissue.
If colonization is successful, the tree ultimately dies, and the next generation of beetles disperses from the parental tree in search of new hosts \cite{13_safranyik_2006}. The detection of infested trees by \textit{Dendroctonus mexicanus} is studied in this paper, which is among the most damaging insects to pine forests in Mexico.  

\indent Emerging bark beetles disperse in a number of ways in search of new hosts, with the majority partaking in short-range dispersal \cite{12_safranyik_1992}. They fly below the forest canopy and attack suitable host trees within a few hundred meters. Hence, identifying attacked trees will help determine the next likely location of infestations and guide beetle management activities (e.g., sanitation, removal, or disposal) to prevent infestations from further spreading \cite{04_hall_2016}. Automated systems can be designed to detect and analyze bark beetle infestations using remote sensing and machine learning (ML), avoiding labor- and cost-intensive efforts of traditionally employed ocular assessments. 
Although satellite and aircraft platforms are widely used at the landscape level, recent research has focused on leveraging UAVs for data collection due to their advantages at the individual tree level (e.g., higher spatial and temporal resolution). Besides, classical ML-based approaches (e.g., random forests (RF) or support vector machines (SVM)) require feature selection which demands prior experience and extensive effort to achieve satisfactory results. Thus, exploiting deep learning (DL)-based models is of interest due to their capacity for ``\textit{learning}" powerful representations and exhibiting good generalization through the discovery of intricate, underlying data patterns. \\
\begin{figure*}[!t]
\centering
\includegraphics[width=0.8\linewidth]{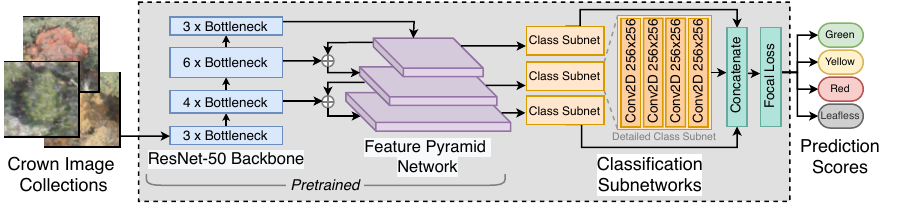}
\vspace{-.3cm}
\caption{\textcolor{black}{An overview of the proposed method. First, the backbone and feature pyramid network (FPN) are initialized using the DeepForest model \cite{DeepForest} trained for tree crown detection. Following that, the network is modified and trained to classify the stages of bark beetle attack.}}
\label{fig:workflow}
\vspace{-0.3cm}
\end{figure*}
\indent Although DL has achieved great success in a wide range of computer vision applications, a few recent studies have attempted to use DL-based models to classify infested trees from UAV-captured images \cite{ref1_DL,ref2_DL,ref3_DL}. Due to the lack of sufficient samples, these works either train customized shallow networks (i.w., six convolutional layers in \cite{ref1_DL,ref3_DL}) or apply transfer learning to models pre-trained on ImageNet \cite{imagenet} (i.e., VGGNet \cite{VGGNet} or ResNet \cite{resnet} in \cite{ref1_DL,ref3_DL}). However, transferring pre-trained models has produced lower results potentially due to the frozen weights that were previously trained for generic object classification purposes (e.g., cats vs. dogs). \\
\indent This paper proposes a DL-based method that exploits modified RetinaNet architecture (pre-trained for tree crown detection) to classify four infestation stages of trees attacked by bark beetles (see Fig.~\ref{fig:workflow}). The simple and efficient architecture allows handling data class imbalance problems and performs well in aerial imagery applications with dense targets. Moreover, the proposed method considers various data augmentation strategies to alleviate the problem of limited number of samples and investigates their effects on network performance. Empirical evaluations demonstrate that the proposed method outperforms the baseline and classical ML methods considerably.
\begin{figure}[b]
\vspace{-0.3cm}
\centering
\includegraphics[width=0.98\linewidth]{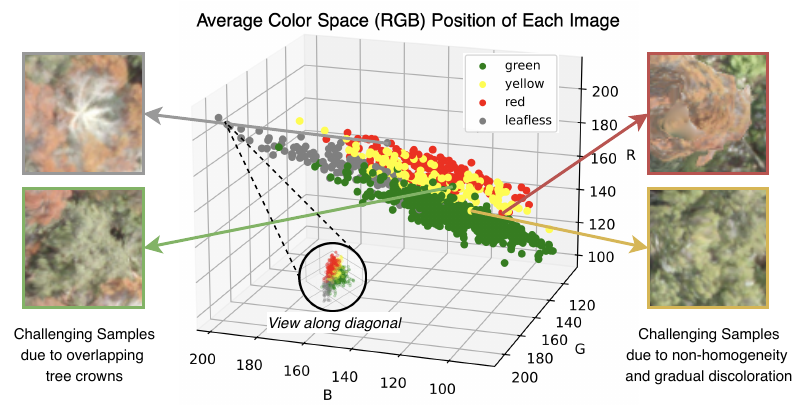}
\vspace{-.3cm}
\caption{RGB color space distribution of bark beetle dataset images. The borders of the highlighted challenging samples indicate their true labels.}
\label{fig:challenge}
\end{figure}

\section{Related Work}
\label{sec:related_works}
This section briefly describes DL-based approaches that seek to detect infested trees by different bark beetle species from UAV captured images (i.e., individual tree level). 
First, the potential of deep neural networks (DNNs) to detect bark beetle outbreaks in fir forests is studied in \cite{ref1_DL}. It employs a two-stage method consisting of a classical image processing-based crown detection and a six-layer convolutional neural network (CNN) for predicting red- and gray-attacked trees by four-eyed fir bark beetles (\textit{Polygraphus proximus Blandford, Coleoptera, Curculionidae}). This method uses RGB images captured by a DJI Phantom 3 Pro quadcopter, and the performance is compared with six well-known CNN models (e.g., VGGNet \cite{VGGNet}, ResNet \cite{resnet}, and DenseNet \cite{DenseNet}). After that, the classification accuracy of infested trees in a temperate forest is investigated in \cite{ref2_DL} by training two shallow CNNs (with three and six convolutional layers) and applying transfer learning to a pre-trained DenseNet-169 \cite{DenseNet}. Despite the availability of multi-spectral images from a DJI Matrice 210 RTK, the best results of this method are obtained using only RGB bands for detecting yellow-attacked trees. Finally, the health statuses of Maries fir trees are evaluated \cite{ref3_DL} by adopting pre-trained CNN models of AlexNet \cite{AlexNet}, SqueezeNet \cite{SE-Res-CVPR}, VGGNet \cite{VGGNet}, ResNet \cite{resnet}, and DenseNet \cite{DenseNet}. Using a DJI Mavic 2 Pro \& DJI Phantom 4 Quadcopter, this method uses RGB images to select treetops in a traditional manner and classify healthy and gray-attacked trees. \\
\indent In contrast, we propose to adapt a state-of-the-art deep network \cite{retinanet} for the classification of bark beetle attacks by exploiting the weights that have been specifically trained for tree crown detection from UAV images and training a shallow subnetwork for discriminating attack stages.

\begin{figure}[b]
\vspace{-0.3cm}
\centering
\includegraphics[width=0.98\linewidth]{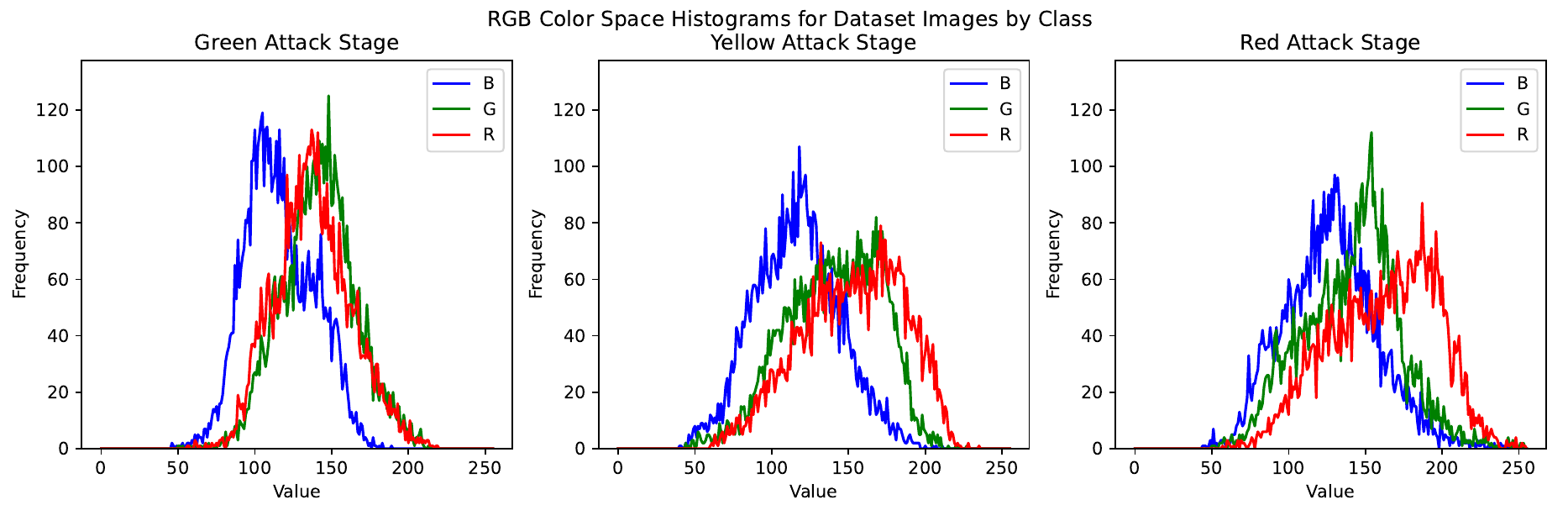}
\vspace{-.3cm}
\caption{Histograms showing RGB color space distribution of the different attack stages with leaves.}
\label{fig:histogram}
\end{figure}

\section{Proposed Method} 
Even though this task seems like a simple color classification, ill-defined attack labels and imbalanced datasets make it more challenging than it appears. For instance, the distribution and some challenging samples are visualized in Fig.~\ref{fig:challenge}, in which green and leafless (needle-less) classes overlap with other classes. 
Also, Fig.~\ref{fig:histogram} shows the RGB color space histograms for each class that reveals similarities between the yellow and red attack stages due to the gradual nature of foliage discoloration. \\
\indent Our proposed method is based on the RetinaNet architecture \cite{retinanet}, which has been successfully used for other remote sensing applications (e.g., \cite{RetinaNet_Application}) owing to its ability to detect dense targets from data with highly imbalanced classes. The proposed RetinaNet-based architecture includes the backbone network (i.e., ResNet-50 \cite{resnet}), feature pyramid network (FPN), classification subnetwork, and focal loss. Although the backbone network seeks to extract multi-scale features, the FPN combines semantically low-resolution features with low-level, high-resolution ones. The classification subnetwork then predicts the category of bark beetle attacks (i.e., green (healthy) tree, yellow-/red-attacked tree, or leafless) using focal loss. This loss function helps to simultaneously handle the inherent similarity of attack classes and limited-sized data by focusing on hard samples and avoiding easy negatives.  \\
\indent As shown in Fig.~\ref{fig:workflow}, the cropped images of tree crowns are normalized according to the mean and standard deviation of the training set images and then fed into the backbone network. The computations are forward propagated through the bottleneck layers, as well as being combined with the layers in the feature pyramid network. Each level of the pyramid feeds its computation to a classification subnetwork, each of which consists of four convolutional layers. Then, the network outputs a score for each attack stage. At last, one-hot encoding is done to get the class prediction for each individual tree\footnote{Bounding box regression subnetwork has been removed considering the available tree crown collections.}.
\textcolor{black}{In contrast to previous studies that train either a shallow network or deep models pre-trained on ImageNet \cite{imagenet}, we exploit a pre-trained deep model (i.e., DeepForest \cite{DeepForest} for tree crown detection) and train the modified network for the classification of attack stages.}
As a result of appropriately initializing network weights with features relevant to tree crowns, the classification subnetwork can learn to differentiate different bark beetle attack stages.
Meanwhile, several different data augmentation strategies are considered to address the imbalance in the dataset. Although it is generally assumed that data augmentation will result in higher performance, we will show that blindly utilizing these techniques can drastically affect classification results for this task.

\section{Empirical Evaluations}
\label{sec:results}
We evaluate the proposed method using the dataset presented in \cite{schaeffer_baseline} that utilized a hexacopter with a Tarot FY680 Pro to capture multiple RGB video sequences of a forested region in Northern Mexico from a top-down perspective. 
Five flights in total were conducted at three different average heights above ground (60m, 90m, and 100m) during three months (June, July, and August).
The individual frames from each flight were combined into five different orthomosaics, and the ground truth information for each tree's center and attack stage were made available \textcolor{black}{(see \cite{schaeffer_baseline} for more details)}. The proposed method is compared with the baseline method \cite{schaeffer_baseline} as well as the most promising SVM, RF, and K-nearest neighbors (KNN) classifiers. Hyperparameter tuning was done in a grid search manner for each classifier.

\subsection{Implementation Details}
We cropped individual tree crowns from the five orthomosaics as square patches and then split them into five separate training, validation, and testing sets, as shown in Table~\ref{tab:dataset}. For each flight, one model is trained and evaluation is performed for each individually and averaged. The proposed networks were trained using the AdamW optimizer for 50 epochs and a batch size of 2 (five models for flights). The training procedure was performed on a Nvidia GeForce RTX 3090 GPU, with each model taking approximately 1.5 hours to train. The dataset was augmented by generating minority class samples using i) random affine warps, ii) vertical/horizontal flips, iii) 90$^{\circ}$/180$^{\circ}$/270$^{\circ}$ rotations, iv) cropping by a factor of 85\%, v) color jittering with random brightness, contrast, \& saturation, and vi) Gaussian blurring with kernel size 5. 
\textcolor{black}{Furthermore, early stopping was considered to avoid overfitting during the training procedure.}

\begin{table}[!tb]
    \vspace{-.1cm}
    \rowcolors{2}{}{gray!10}
    \caption{Dataset Distribution for Each Flight by Class.}
    \vspace{-.2cm}
    \label{tab:dataset}
    \centering
    \begin{tabular}{c | c c c c c}
        \toprule
        \textcolor{black}{Subsets of Samples}  & Jun60 & Jul90 & Jul100 & Aug90 & Aug100 \\
        \midrule
        Green Trees    & 68 & 81 & 103 & 141 & 98\\
        Yellow Trees   & 34 & 19 & 28  & 45  & 49\\
        Red Trees     & 24 & 26 & 48  & 52  & 48\\
        Leafless Trees & 25 & 28 & 26  & 33  & 25\\
        \midrule
        Train   & 128 & 130 & 174 & 230 & 187\\
        \textcolor{black}{Augmented Train} & 232 & 276 & 352 & 480 & 332\\
        Validation     & 7   & 7   & 10  & 13  & 11\\
        Test    & 16  & 17  & 21  & 28  & 22\\
        \bottomrule
    \end{tabular}
    \vspace{-0.2cm}
\end{table}
\begin{table}[!b]
    \vspace{-0.4cm}
    \rowcolors{2}{}{gray!10}
    \caption{Comparison of Classification Accuracy from Various Models.}
    \vspace{-.2cm}
    \label{tab:results}
    \centering
    \begin{tabular}{c c | c}
        \toprule
        & Model & Average Accuracy ($\uparrow$)\\
        \midrule
         & SVM & 53.10\%\\
         & KNN & 53.10\%\\
         & RF & 40.24\%\\
         & Baseline \cite{schaeffer_baseline} (Best result) & 89\%\\
         & \textbf{Ours (with warping)} & \textbf{98.95\%}\\
        \midrule
        \global\let\CT@@do@color\relax  \multirow{6}{*}{\rotatebox[origin=c]{90}{Ablation Study}} & \global\let\CT@@do@color\oriCT@@do@color Ours (without augmentation) & 97.69\%\\
        \global\let\CT@@do@color\relax  & \global\let\CT@@do@color\oriCT@@do@color Ours (with cropping) & 96.29\%\\
        \global\let\CT@@do@color\relax  & \global\let\CT@@do@color\oriCT@@do@color Ours (with flips) & 94.74\%\\
        \global\let\CT@@do@color\relax  & \global\let\CT@@do@color\oriCT@@do@color Ours (with blurring) & 92.23\%\\
        \global\let\CT@@do@color\relax  & \global\let\CT@@do@color\oriCT@@do@color Ours (with rotation) & 84.71\%\\
        \global\let\CT@@do@color\relax  & \global\let\CT@@do@color\oriCT@@do@color Ours (with color jittering) & 83.90\%\\
        \bottomrule
    \end{tabular}
    \vspace{-0.3cm}
\end{table}
\subsection{Experimental Results}
\begin{figure}[!t]
\centering
\includegraphics[width=0.98\linewidth]{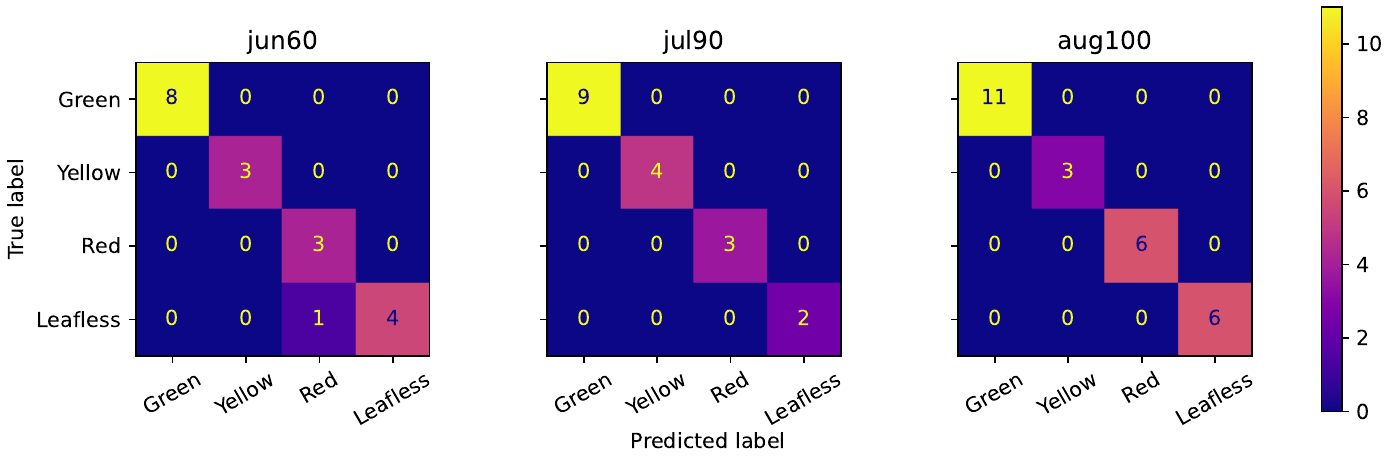}
\vspace{-.2cm}
\caption{Confusion matrices for our best performing network (Jul100m and Aug90m omitted since there were no misclassifications).}
\label{fig:confusion_matrix}
\vspace{-0.2cm}
\end{figure}

The experimental comparison of the proposed method (including the best model with affine warping data augmentation) with the baseline and best performing models for classical ML methods is shown in Table~\ref{tab:results}. According to the results, the proposed method considerably outperforms (by 9.9\% (\& 7.6\%) with (\& without) data augmentation in average accuracy) the cellular automaton baseline method. 
Also, the classical ML methods have achieved significantly lower accuracy than our method, which can be explained by the ill-defined separation between classes in the RGB color space, as shown in Fig.~\ref{fig:challenge}. The confusion matrices for the challenging flights are shown in Fig.~\ref{fig:confusion_matrix}. Accordingly, the proposed method has no misclassifications for four of the flights, and only one leafless image is incorrectly predicted as red in the June 60m flight due to the considerable overlap from nearby red attack stage trees. 
Since classic ML methods rely on manual feature selection, applying them directly to raw data results in poor performance. However, the proposed DL-based method can automatically learn the most relevant and robust features from the dataset, enabling it to perform significantly better.

\subsection{Ablation Study} 
\label{subsec:ablation}
To assess the effectiveness of data augmentation, various probabilistic augmentation strategies are applied. In each strategy, additional samples belonging to the red, yellow, and leafless classes are randomly generated to obtain the same number as the green samples (i.e., balancing the dataset). The classification results for models trained on each strategy are shown in Table~\ref{tab:results}.
Accordingly, affine warping is found to be the most effective strategy considering tree crowns are not always circular. This strategy changes the apparent geometry of the trees, promoting more diversity in the dataset. Also, it accounts for angular variation in the UAV during data collection.
Color jittering unsurprisingly leads to the most performance degradation (major effect based on visual symptoms of trees). These results can further be explained using the t-SNE visualizations in Fig.~\ref{fig:tsne}. The middle and left plots display similar separations in the dataset, indicating that warping adds minority class samples without adversely impacting the separation of the classes. On the other hand, the right plot is obtained from the color-jittered dataset, and significantly more overlap between the classes can be observed (e.g., in the bottom right corner). The other augmentation strategies do not improve performance either.

\section{Conclusion}
A DL-based approach was proposed to classify bark beetle-infested trees in this work. Based on the RetinaNet architecture, the proposed method simultaneously trains a shallow subnetwork and exploits a deep network initialized with weights trained to detect tree crowns from UAV images.
To overcome the data imbalance problem, different data augmentation strategies were investigated and affine warping is found the most effective for this purpose. Despite the challenges of inter-class overlap and intra-class non-homogeneity in the dataset, the proposed method achieves an average accuracy of 98.95\%, thereby significantly outperforming the baseline method. 
\textcolor{black}{Future work involves accurate detection of tree crowns from aerial orthomosaic images and discrimination of non-beetle factors that may result in similar foliage discoloration (e.g., drought).}

\newcommand{\imgWidth}{0.32\linewidth}
\begin{figure}[!t]
\vspace{-0.3cm}
\centering
\subfloat[No augmentation]{\includegraphics[width=\imgWidth]{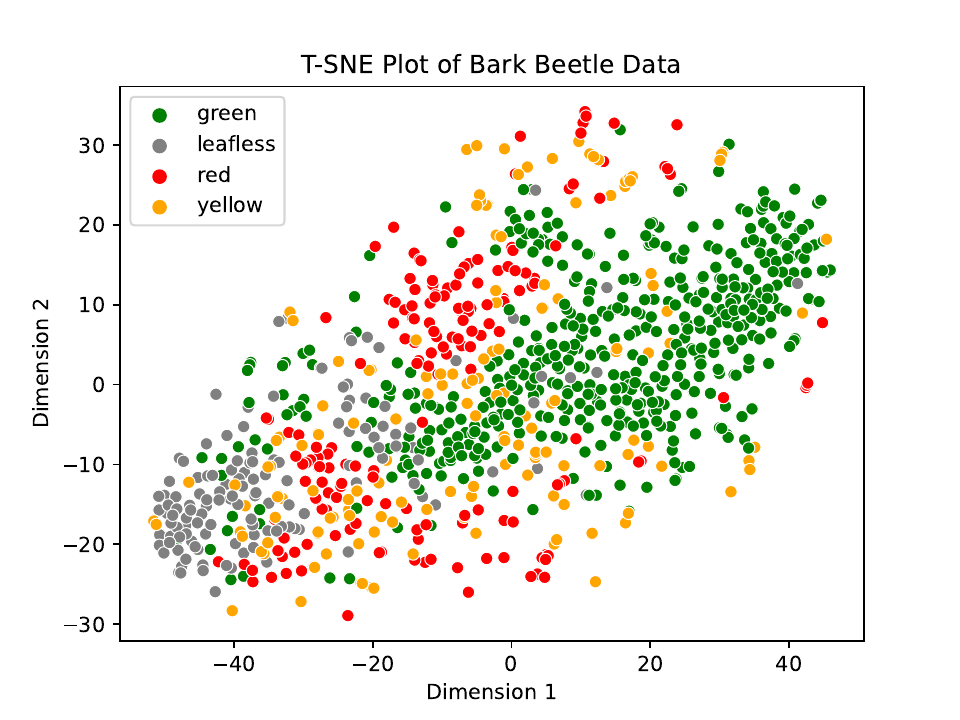}%
\label{subfig:tsne_none}}
\hfil
\subfloat[Affine warping]{\includegraphics[width=\imgWidth]{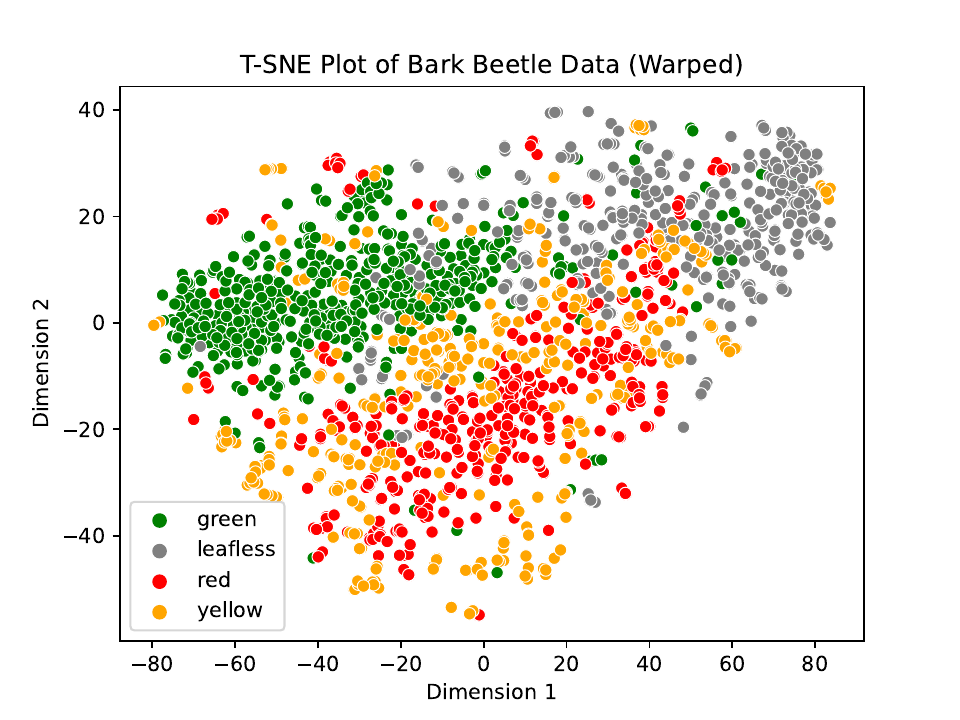}%
\label{subfig:tsne_warp}}
\hfil
\subfloat[Color jittering]{\includegraphics[width=\imgWidth]{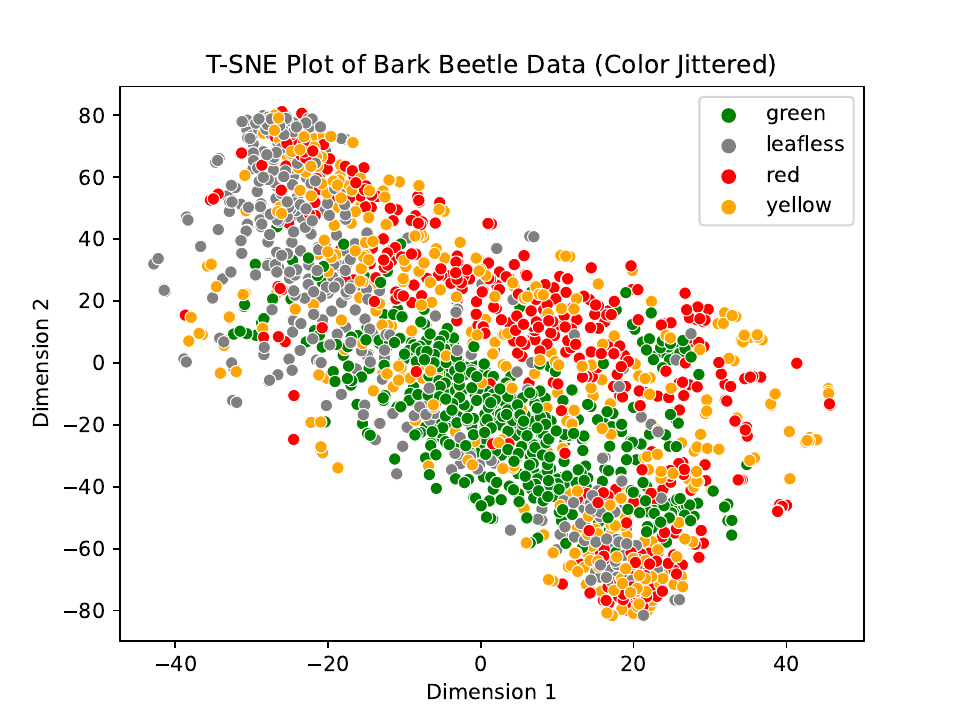}%
\label{subfig:tsne_jitter}}
\caption{T-SNE visualization of dataset with different augmentations.}
\label{fig:tsne}
\vspace{-0.2cm}
\end{figure}

\section*{Acknowledgments}
\noindent This work was supported in-part by the Federal-Provincial Mountain Pine Beetle Research Partnership program.

\bibliographystyle{./IEEEtran}
\bibliography{IEEEabrv,refs}

\end{document}